# Theme-weighted Ranking of Keywords from Text Documents using Phrase Embeddings

Debanjan Mahata
University of Arkansas at Little Rock
dxmahata@ualr.edu

John Kuriakose
Infosys Ltd.
John_Kuriakose@infosys.com

Rajiv Ratn Shah
IIIT-Delhi
rajivratn@iiitd.ac.in

Roger Zimmermann
National University of Singapore
rogerz@comp.nus.edu.sg

John R. Talburt
University of Arkansas at Little Rock
jrtalburt@ualr.edu

*Abstract*—Keyword extraction is a fundamental task in natural language processing that facilitates mapping of documents to a concise set of representative single and multi-word phrases. Keywords from text documents are primarily extracted using supervised and unsupervised approaches. In this paper, we present an unsupervised technique that uses a combination of theme-weighted personalized PageRank algorithm and neural phrase embeddings for extracting and ranking keywords. We also introduce an efficient way of processing text documents and training phrase embeddings using existing techniques. We share an evaluation dataset derived from an existing dataset that is used for choosing the underlying embedding model. The evaluations for ranked keyword extraction are performed on two benchmark datasets comprising of short abstracts (Inspec), and long scientific papers (SemEval 2010), and is shown to produce results better than the state-of-the-art systems.

## I. BACKGROUND AND INTRODUCTION

Keywords are single or multi-word linguistic units that represent the salient aspects of a document. In this paper, we use the term keyword uniformly to represent both single and multi-word phrases. Keywords are useful in many tasks such as indexing documents [1], summarization [2], clustering [3], ontology creation [4], classification [5], auto-tagging [6] and visualization of text [7]. Due to its widespread use, keyword extraction has received a lot of attention from researchers.

In order to push the state-of-the-art performances of keyword extraction systems, the research community has been hosting shared tasks like SemeEval 2010 Task 5 [8] and SemEval 2017 Task 10 [9]. However, the task is far from solved and the performances of the present systems are worse in comparison to many other NLP tasks [10]. Some of the major challenges are varied length of the documents to be processed, their structural inconsistency and developing strategies that can perform well in different domains [11].

Methods for automatic keyword extraction are mainly divided into two categories: *supervised* and *unsupervised*. Supervised methods approach the problem of keyword extraction as a binary classification problem [11], whereas the unsupervised methods are mostly based on TFIDF, clustering, and graph-based ranking [12]. On presence of domain-specific data, supervised methods have shown better performance than the unsupervised methods. The unsupervised methods have the advantage of not requiring any training data and can produce results in any domain. However, the assumptions of unsupervised methods do not hold for every type of document.

With recent advancements in deep learning techniques applied to natural language processing, the latest trend is to represent words as dense real-valued vectors obtained by training a shallow neural network on a fixed vocabulary. These distributional representation of words are popularly known as word embeddings. These neural representations have been shown to equal or outperform other methods (e.g. LSA, SVD) [13]. The vector representation of words, are supposed to preserve the semantic and syntactic similarities between them. They have been shown to be useful for several natural language processing (NLP) tasks, like part-of-speech tagging, chunking, named entity recognition, semantic role labeling, syntactic parsing, and speech processing, among others [14]. Some of the most popular approaches for generating word embeddings are Word2Vec [15], Glove [16] and Fasttext [17].

Word embeddings have already shown promising results in the process of keyword extraction from scientific articles [18], [19]. However, Wang *et al.* did not use domain-specific word embeddings and had suggested that training them might lead to improvements. This motivated us to experiment with domain-specific embeddings on scientific articles. In this work, we represent candidate keywords extracted from a scientific article by domain-specific phrase embeddings and rank them using a *theme-weighted* PageRank algorithm [20], such that the *thematic weight* of a candidate keyword indicate how similar it is to the *thematic representation* of the article, which is also constructed using the embeddings. To our knowledge, using multi-word phrase embeddings for constructing *thematic representation* of a given document and to assign *thematic weights* to phrases have not been used for ranked keyword extraction, and this work is the first attempt to do so. Table 1. shows keywords extracted by using our method from a sample research article abstract.

We not only aim at extracting the keywords that are statistically relevant for the given document, as mostly attempted by previous studies [11], but also to rightly identify meaningful phrases that are most semantically related to the main theme of the document. We are motivated by the following properties

**Title:** Identification of states of complex systems with estimation of admissible measurement errors on the basis of fuzzy information.
**Abstract:** The problem of identification of states of complex systems on the basis of fuzzy values of informative attributes is considered. Some estimates of a maximally admissible degree of measurement error are obtained that make it possible, using the apparatus of fuzzy set theory, to correctly identify the current state of a system.
**Automatically identified keywords:** *complex systems, fuzzy information, admissible measurement errors, fuzzy values, informative attributes measurement error, maximally admissible degree, fuzzy set theory*
**Manually assigned keywords:** *complex system state identification, admissible measurement errors, informative attributes, measurement errors fuzzy set theory*

TABLE I: Keywords extracted by using our method from a sample research article abstract.

for the top-K keywords that we select, as mentioned in [21].

- *Understandability* - The ranked keywords should be understandable to readers, which is made possible by chosing grammatically correct and meaningful phrases that possess the characteristic of high readability by humans (Section II-A). For example, the phrase *scientific articles* has more understandability than the individual words *scientific* and *articles*.
- *Relevancy* - The top-K keywords should be semantically related to the central theme of the document (Section II-D).
- *Good Coverage* - The keywords should cover all the major topics discussed in the document (Section II-F).

This intrigued us to look at the route of training phrase embeddings as presented in this paper, rather than first training embedding models for unigram words and then combining their dense vector representations to obtain similar representations for multi-word phrases.

The major contributions of this paper are,

- *Efficient processing of text for training neural phrase embeddings using existing techniques for training word embeddings.*
- *Thematic representation of text documents using phrase embeddings and assignment of thematic weights to candidate keywords.*
- *Theme-weighted personalized PageRank algorithm for automatic ranking of candidate keywords extracted from a text document.*

Next, we give a detailed description of our methodology.

## II. METHODOLOGY

In our framework we primarily use three steps: *candidate selection*, *candidate scoring*, and *candidate ranking*. All the steps depend on a phrase embedding model that we train, and the choice of our text processing steps. We explain them next and give a detailed description of their implementations.

### A. Text Processing

It has been shown [15], that the presence of multi-word phrases intermixed with unigram words increases the performance and accruacy of the embedding models trained using techniques such as *Word2Vec*. However, in our framework we take a different approach in detecting meaningful and cohesive chunk of phrases while preparing the text samples for training. Instead of relying on measures considering how often two or more words co-occur with each other, we rely on already trained dependency parsing and named entity extraction models provided by *Spacy*[1]. We split a text document into sentences, tokenize a sentence into unigram tokens, as well as identify noun phrases and named entities from it. During this process if a named entity is detected at a particular offset in the sentence then a noun phrase appearing at the same offset is not considered.

We take steps in cleaning the individual single word and multi-word tokens that we obtain. Specifically, we filter out the following tokens.

- Noun phrases and named entities that are fully numeric.
- Named entities that belong to the following categories are filtered out : DATE, TIME, PERCENT, MONEY, QUANTITY, ORDINAL, CARDINAL. Refer, Spacy's named entity documentation[2] for details of the tags.
- Standard stopwords are removed.
- Punctuations are removed except '-'.

We also take steps to clean leading and ending tokens of a multi-word noun phrase and named entity.

- Common adjectives and reporting verbs are removed if they occur as the first or last token of a noun phrase/named entity.
- Determiners are removed from the first token of a noun phrase/named entity.
- First or last tokens of noun phrases/named entities belonging to following parts of speech: INTJ Interjection, AUX Auxiliary, CCONJ Coordinating Conjunction, ADP Adposition, DET Interjection, NUM Numeral, PART Particle, PRON Pronoun, SCONJ Subordinating Conjunction, PUNCT Punctutation, SYM Symbol, X Other, are removed. For a detailed reference of each of these POS tags please refer Spacy's documentation[3].
- Starting and ending tokens of a noun phrase/named entity is removed if they belong to a standard list of english stopwords.
- Starting and ending tokens of a noun phrase/named entity is removed if they belong to a standard list of english functional words.

Apart from relying on Spacy's parser we use hand crafted regexes for cleaning the final list of tokens obtained after the above data cleaning steps. The regexes are obtained from Textacy[4], which is a text mining library built on top of *Spacy*.

- Get rid of leading/trailing junk characters.
- Handle dangling/backwards parentheses. We don't allow '(' or ')' to appear without the other.
- Handle oddly separated hyphenated words.
- Handle oddly separated apostrophe'd words.
- Normalize whitespace.

---
[1]https://spacy.io
[2]https://spacy.io/usage/linguistic-features#section-named-entities
[3]https://spacy.io/api/annotation#pos-tagging
[4]https://github.com/chartbeat-labs/textacy/blob/master/textacy/text_utils.py

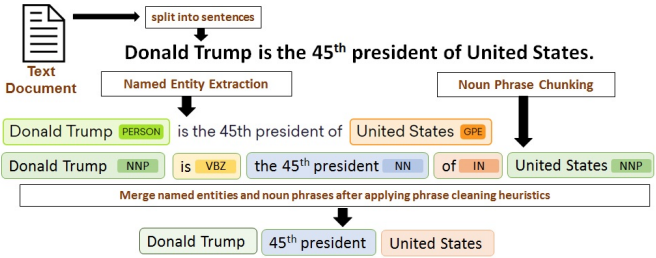

Fig. 1: Text processing pipeline

The resultant unigram tokens and multi-word phrases are merged in the order they appeared in the original sentence. Figure 1, shows an example of how the text processing pipeline works on an example sentence for preparing the training samples that act as an input to the embedding algorithm.

### B. Embedding Model Selection

In this section, we build on our previous work [22] and train our embeddings using popular techniques. For implementing our framework, a phrase embedding model is needed that can be used for constructing semantically aware representations of phrases, sentences and textual content that forms as a basis for calculating similarities. We are mainly interested in capturing three kinds of similarities *phrase-phrase*, *sentence-sentence* and *phrase-sentence*, respectively. Although, we don't use phrase-phrase similarity, yet it is fundamental towards understanding the quality of the trained models and also acts as a building block for the other types of similarity calculations. In order to evaluate the models for the above criterias, we needed an evaluation dataset. We created three datasets from an exsiting dataset[5], originally developed for evaluating document representations that capture document similarities [23].

*1) Evaluation Dataset:* The first dataset consists of 106 triplets, which is a subset of manually curated triplets provided in the original dataset. It consists of three phrases in a triplet for evaluating *phrase-phrase* similarity, where the first phrase is semantically closer to the second phrase than it is closer to the third phrase. For example, *Deep learning* is closer to *Machine learning* than *Computer network* or *September* is closer to *October* than *June*. The second dataset consists of 6247 triplets, which is also a subset of automatically generated triplets shared in the original dataset. The triplets are phrases that are titles of Wikipedia articles and are also used for evaluating *phrase-phrase* similarity. The first phrase is supposed to be closer to the second than the third, on the basis, that the first and second phrases are titles of articles belonging to the same category in Wikipedia, unlike the third phrase, which is the title of an article from Wikipedia that belongs to a different category. Contrary to the original dataset that uses the full content of the Wikipedia articles mapped to these triplets for evaluating document similarities, we only use the title phrases.

The third dataset consists of 6,353 triplets that we derive from the first two datasets and original articles automatically collected from Wikipedia. The triplets are *phrase-sentence* combinations intended for evaluating *phrase-sentence* and *sentence-sentence* similarities. We combine the triplets from first two datasets and collect all the Wikipedia articles mapped to them using a crawler. Each part of the triplet consists of a combination of a phrase, which is the title of a Wikipedia article, and the first sentence of the article that mentions that phrase. The first phrase is supposed to be semantically more similar to the sentence associated with it than the sentence associated with the second or the third phrase. For example, *Deep learning* is closer to *"Deep learning (also known as deep structured learning, hierarchical learning or deep machine learning) is a class of machine learning algorithms that: use a cascade of many layers of nonlinear processing units for feature extraction and transformation"*, than *"A computer network or data network is a telecommunications network which allows nodes to share resources"*, which is a sentence about *computer network*. Also, the first sentence is closer to the second sentence (*Machine learning is the subfield of computer science that, according to Arthur Samuel in 1959, gives "computers the ability to learn without being explicitly programmed."*) than the third sentence. Additionally, we automatically collect full content of 17,326 Wikipedia articles mapped to the titles of the triplets provided in the original dataset, which we use for training different configurations of phrase embedding models that are further used for carrying out the similarity evaluation tasks. The entire evaluation dataset is publicly shared[6].

*2) Training:* We process the text of the Wikipedia articles as explained in Section II-A and prepare the dataset for training phrase embeddings. In order to select the model configurations that would best capture the underlying similarities between different textual units, we train phrase embedding models using *skipgram* and *continuous bag of words* schemes as implemented in *Word2Vec*[7] and *Fasttext*[8] toolkits. The vocabulary size of all our models is 3,000,664 unique phrases. In this work, we use *negative sampling* for all the schemes, with the number of negative samples fixed to 5. We also fix the size of the *context window* to 5 and number of *epochs* to 10. For producing the vector representations of larger blocks of text, like sentences, we average out the vectors of individual phrases extracted from it. The accuracies of the trained models for different dimensions (10 - 1000) on three different similarity tasks *phrase-phrase*, *sentence-sentence* and *phrase-sentence* are reported in Tables II, III and IV, respectively.

The models that we train in this section is not used directly for the downstream process of ranked keyword extraction. We perform these training and evaluations in order to narrow down to an optimal set of parameter configurations and scheme for training the main phrase embedding model that we use for implementing our framework. We are interested to try out many other tools and configurations and study the effects on the quality of the phrase embeddings, as a part of our future work. We believe that the dataset developed and shared in this paper would allow us and the community at large for carrying out such studies. From the table we can clearly see that the models trained using *Fasttext* performs better than the models trained using *Word2Vec* on all the three tasks. After analyzing the accuracies of the models, we decided to use 100 dimensional phrase embeddings trained using *skipgram* method and *negative sampling*, as implemented in the *Fasttext* toolkit[9]. In the future sections these configurations should be assumed for the underlying phrase embedding model.

---

[5]http://cs.stanford.edu/ quocle/triplets-data.tar.gz

[6]www.example.com
[7]https://radimrehurek.com/gensim/models/word2vec.html
[8]https://github.com/facebookresearch/fastText

| Model | Dataset | 10 | 25 | 50 | 75 | 100 | 500 | 1000 |
|---|---|---|---|---|---|---|---|---|
| Word2Vec Skipgram | manual triples | 72.65% | 84.60% | 83.52% | 83.52% | 87.86% | 87.86% | 84.60% |
| | auto triples | 62.48% | 64.98% | 64.89% | 64.60% | 64.98% | 66.29% | 64.77% |
| Word2Vec CBOW | manual triples | 65.04% | 80.26% | 83.52% | 81.34% | 80.26% | 81.34% | 79.17% |
| | auto triples | 60.70% | 62.10% | 60.11% | 61.50% | 61.42% | 61.08% | 61.50% |
| Fasttext Skipgram | manual triples | 79.17% | 88.95% | 92.17% | 94.39% | 90.04% | 86.95% | 90.04% |
| | auto triples | 64.85% | 67.05% | 67.65% | 67.18% | 68.28% | 64.12% | 67.90% |
| Fasttext CBOW | manual triples | 73.73% | 82.43% | 88.95% | 85.69% | 90.04% | 86.95% | 83.52% |
| | auto triples | 62.14% | 65.32% | 64.89% | 65.19% | 64.68% | 64.12% | 63.41% |

TABLE II: Accuracies of phrase embedding models for phrase-phrase similarity task.

| Model | 10 | 25 | 50 | 75 | 100 | 500 | 1000 |
|---|---|---|---|---|---|---|---|
| Word2Vec Skipgram | 63.00% | 65.63% | 66.56% | 65.74% | 66.36% | 66.25% | 65.98% |
| Word2Vec CBOW | 60.52% | 62.69% | 63.78% | 63.93% | 63.85% | 64.03% | 63.79% |
| Fasttext Skipgram | 65.59% | 69.05% | 70.04% | 70.51% | 70.94% | 70.95% | 71.03% |
| Fasttext CBOW | 63.65% | 66.67% | 67.79% | 67.92% | 67.77% | 68.01% | 67.40% |

TABLE III: Accuracies of phrase embedding models for sentence-sentence similarity tasks.

### C. Training Embedding Model

Since this work deals with the domain of scientific articles we train our phrase embedding model on a collection of more than million scientific documents. We collect 1,147,000 scientific abstracts related to different areas from arxiv.org[10]. For collecting data we use the API provided by arxiv.org that allows bulk access of the articles uploaded to their portal. A distribution of scientific abstracts from different topics present in our dataset is shown in Fig 2. We also add the scientific documents present in the benchmark datasets (Sections III). After processing the text of 1,149,244 scientific documents as mentioned in Section II-A, we train our model using the configurations chosen in the previous section. Table V. shows top 5 similar phrases for five different phrases as produced by the trained phrase embedding model. We use this model as the underlying phrase embedding model.

[9]https://fasttext.cc/

| Model | 10 | 25 | 50 | 75 | 100 | 500 | 1000 |
|---|---|---|---|---|---|---|---|
| Word2Vec Skipgram | 73.07% | 76.40% | 76.75% | 76.98% | 76.96% | 76.75% | 76.58% |
| Word2Vec CBOW | 61.27% | 64.67% | 66.37% | 67.37% | 68.58% | 68.44% | 69.11% |
| Fasttext Skipgram | 82.89% | 90.08% | 93.25% | 94.36% | 94.71% | 96.18% | 96.27% |
| Fasttext CBOW | 78.05% | 85.89% | 89.73% | 91.01% | 92.22% | 93.99% | 93.92% |

TABLE IV: Accuracies of phrase embedding models for phrase-sentence similarity task.

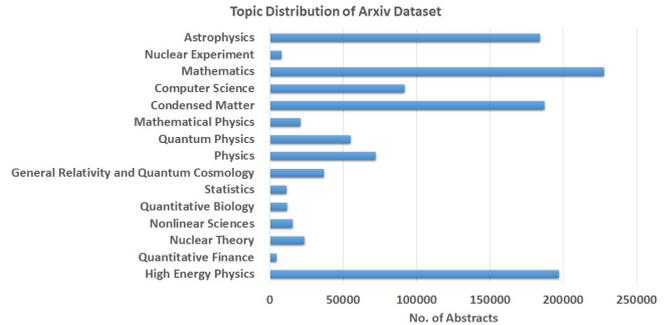

Fig. 2: Frequency distribution of topics in the arxiv dataset used for training phrase embeddings.

| Phrase | Top 5 Similar Phrases |
|---|---|
| convolutional_neural_network | cnn, feature_representations, deep_convolutional_neural_network, deep_neural_network, scene_recognition |
| dark_matter | dm, dark_matter_particle, non-baryonic_dark_matter, dark_energy, self-interacting_dark_matter |
| natural_language_processing | nlp, language_processing, machine_translation, named_entity_recognition, sense_disambiguation |
| rnn | blstm, long_short-term_memory, lstms, handwritten_documents, recurrent_neural_network, lstm |
| svm | support_vector_machine, support_vector_machines, random_forest, svms, naive_bayes |

TABLE V: Top 5 similar phrases to a given phrase as produced by the phrase embedding model trained on the arxiv dataset using Fasttext (SkipGram).

### D. Candidate Selection

This step aids in chosing candidate keywords from the set of all phrases that can be extracted from a document, and is commonly used in most of the automated ranked keyword extraction systems. Not all the phrases are considered as candidates. Generally, unwanted and noisy phrases are eliminated in this process by using different heuristics. We use Spacy for splitting a document into sentences and to extract noun phrases and named entities as described in Section II-A. As an output of this step we get a set of unique phrases ($C_{d_i} = \{c_1, c_2, ..., c_n\}_{d_i}$) for a document $d_i$ to be used later for scoring and ranking in the next two steps.

[10]http://arxiv.org

### E. Candidate Scoring

In this step we assign a *theme vector* ($\hat{\tau_{d_i}}$) to a document ($d_i$). The *theme vector* can be tuned according to the type of documents that are being processed and the type of keywords that we want to get in our final results. In this work, we extract a *theme excerpt* from a given document and further extract a unique set of *thematic phrases* comprising of named entities, noun phrases and unigram words ($T_{d_i} = \{t_1, t_2, ..., t_m\}_{d_i}$) from it. For the *Inspec* dataset we use the first sentence of the document that contains the title of the abstract, and for the *SemEval* dataset we use the title and the first ten sentences extracted from the beginning of the document, as the *theme excerpts*, respectively. The first ten sentences of a document from the *SemEval* dataset essentially captures the abstract and sometimes first few sentences of the introduction of a scientific article. We get the vector representation ($\hat{t_j}$) of each *thematic phrase* extracted from the *theme excerpt* using the phrase embedding model that we trained and perform vector addition in order to get the final *theme vector* ($\hat{\tau_{d_i}} = \sum_{j=1}^{m} \hat{t_j}$) of the document. The phrase embedding model is then used to get the vector representation ($\hat{c_k}; k \in \{1...n\}$) for each candidate keyword in $C_{d_i}$.

We calculate the *cosine distance* between the *theme vector* ($\hat{\tau_{d_i}}$) and vector for each candidate keyword ($\hat{c_k}$) and assign a score ($\kappa(\hat{x}, \hat{y}) \to [0, 1]$) to each candidate, with 1 indicating a complete similarity with the *theme vector* and 0 indicating a complete dissimilarity. For getting the final *thematic weight* ($w_{c_j}^{d_i}$) for each candidate w.r.t given document ($d_i$) the candidate scores are scaled again between 0 and 1 with a score of 1 assigned to the candidate semantically closest to the main theme of the document and 0 to the farthest.

### F. Candidate Ranking

In order to perform final ranking of the candidate keywords we use weighted personalized PageRank algorithm. A directed graph $G_{d_i}$ is constructed for a given document ($d_i$) with $C_{d_i}$ as the vertices and $E_{d_i}$ as the edges connecting two candidate keywords if they co-occur within a window size of 5, before performing the text processing steps. The edges are bidirectional. Weights are calculated for the edges using the semantic similarity between the candidate keywords obtained from the phrase embedding model and their frequency of co-occurrence, as used by Wang *et al.* [18], and shown in equation 1. We use the combination of *cosine distance* ($\frac{1}{1-cosine(c_j^{d_i}, c_k^{d_i})}$) and *Point-wise Mutual Information* ($PMI(c_j^{d_i}, c_k^{d_i})$) for calculating $semantic(c_j^{d_i}, c_k^{d_i})$ and $cooccur(c_j^{d_i}, c_k^{d_i})$, respectively. The main intuition behind calculating semantic relatedness by using a phrase embedding model is to capture how well two phrases are related to each other in general. Whereas, the co-occurrence score captures the local relationship between the phrases within the context of the given document.

$$sr(c_j^{d_i}, c_k^{d_i}) = semantic(c_j^{d_i}, c_k^{d_i}) \times cooccur(c_j^{d_i}, c_k^{d_i}) \quad (1)$$

Given graph $G$, if $\varepsilon(c_j^{d_i})$ be the set of all edges incident on the vertex $c_j^{d_i}$, and $w_{c_j}^{d_i}$ is the *thematic weight* of $c_j^{d_i}$ as calculated in the *candidate scoring* step, then the final PageRank score $R(c_j^{d_i})$ of a candidate keyword $c_j^{d_i}$ is calculated using equation 2, where $d = 0.85$ is the *damping factor* and $out(c_k^{d_i})$ is the out-degree of the vertex $c_k^{d_i}$.

$$R(c_j^{d_i}) = (1-d)w_{c_j}^{d_i} + d \times \sum_{c_k^{d_i} \in \varepsilon(c_j^{d_i})} \left(\frac{sr(c_j^{d_i}, c_k^{d_i})}{\left|out(c_k^{d_i})\right|}\right) R(c_k^{d_i}) \quad (2)$$

Next, we evaluate the performance of our system on two different benchmark datasets and compare the results against some state-of-the-art systems known to perform well on these datasets.

## III. EXPERIMENTS AND RESULTS

The final ranked keywords obtained using our methodology as described in the previous section is evaluated on the popular *Inspec* and *SemEval 2010* datasets. The Inspec dataset [24] is composed of 2000 abstracts of scientific articles divided into sets of 1000, 500, and 500, as training, validation and test datasets respectively. Each document has two lists of keywords assigned by humans - *controlled*, which are assigned by the authors, and *uncontrolled*, which are freely assigned by the readers. The controlled keywords are mostly abstractive, whereas the uncontrolled ones are mostly extractive [18]. The *Semeval 2010* dataset [8] consists of 284 full length ACM articles divided into a test set of size 100, training set of size 144 and trial set of size 40. Each article has two sets of human assigned keywords: the *author-assigned* and *reader-assigned* ones, equivalent to the *controlled* and *uncontrolled* categories, respectively of the *Inspec* dataset. We only use the test datasets for our evaluations and combine the annotated *controlled* and *uncontrolled* keywords.

The ranked keywords are evaluated using exact match evaluation metric as used in SemEval 2010 Task 5. We match the keywords in the annotated documents in the benchmark datasets with those generated by our method, and calculate micro-averaged precision, recall and F-score ($\beta = 1$), respectively. In the evaluation, we check the performance over the top 5, 10 and 15 candidates returned by our system. The performance of our system on the metrics is shown in Table VI. Tables VII and VIII shows a comparison of our system with some of the state-of-the-art systems giving best performances on the *Inspec* and *SemEval 2010* datasets, respectively.

## IV. CONCLUSION AND FUTURE WORK

In this paper, we proposed a framework for automatic extraction and ranking of keywords. We showed an efficient way of training phrase embeddings using existing techniques, and showed its effectiveness in constructing thematic representation of text documents and assigning thematic weights to candidate keywords. We also introduced theme-weighted personalized PageRank to rank the candidate keywords. Experimental evaluations confirm that our proposed system produces state-of-the-art results on benchmark datasets. In the future, we plan to use our methodology in multimodal datasets for extraction and generation of keywords for scientific images associated with the documents, for the purpose of automatic tagging and indexing. We would also like to explore other embedding methods and other ranking strategies.

|  | Micro Avg. Precision@5 | Micro Avg. Recall@5 | Micro Avg. F1@5 | Micro Avg. Precision@10 | Micro Avg. Recall@10 | Micro Avg. F1@10 | Micro Avg. Precision@15 | Micro Avg Recall@15 | Micro Avg F1@15 |
|---|---|---|---|---|---|---|---|---|---|
| Inspec | 61.78 | 25.67 | 36.27 | 57.58 | 42.09 | 48.63 | 55.90 | 50.06 | 52.82 |
| SemEval | 41 | 14.37 | 21.28 | 35.29 | 24.67 | 29.04 | 34.39 | 32.48 | 33.41 |

TABLE VI: Performance of our system over combined *controlled* and *uncontrolled* annotated keyphrases for *Inspec* and *SemEval 2010* datasets.

| Inspec (Combined) | Key2Vec | Wang et al., 2015 | Liu et al., 2010 | SGRank | TopicRank |
|---|---|---|---|---|---|
| Micro Avg. F1@10 | 48.63 | 44.7 | 45.7 | 33.95 | 27.9 |

TABLE VII: Comparison of *Our System* with some state-of-the-art systems [25], [18], [21], [26], for F1@10 on *Inspec* dataset.

| SemEval 2010 (Combined) | Our System | SGRank | HUMB | TopicRank |
|---|---|---|---|---|
| Avg F1@10 | 29.04 | 26.07 | 22.50% | 12.1 |

TABLE VIII: Comparison of *Key2Vec* with some state-of-the-art systems for F1@10 on *SemEval 2010* dataset [27].

*Acknowledgement*

This research was supported in part by the National Natural Science Foundation of China under Grant no. 61472266, by the National University of Singapore (Suzhou) Research Institute, 377 Lin Quan Street, Suzhou Industrial Park, Jiang Su, People's Republic of China, 215123, and Infosys Ltd.